\documentclass[conference, compsoc, twoside]{IEEEtran}
\usepackage{fancyhdr}
\ifCLASSOPTIONcompsoc
  \usepackage[nocompress]{cite}
\else
  \usepackage{cite}
\fi
\ifCLASSINFOpdf
\else
\fi

\usepackage{url}
\usepackage{multirow}
\usepackage{array}
\usepackage{svg}

\begin{document}
\IEEEoverridecommandlockouts
%

\title{Agree to Disagree: Improving Disagreement Detection with Dual GRUs}

\author{\IEEEauthorblockN{Sushant Hiray\IEEEauthorrefmark{1}\IEEEauthorrefmark{2},
Venkatesh Duppada\IEEEauthorrefmark{1}\IEEEauthorrefmark{3}}
\IEEEauthorblockA{Seernet Technologies, LLC \\
Email: \IEEEauthorrefmark{2} sushant.hiray@seernet.io,
\IEEEauthorrefmark{3} venkatesh.duppada@seernet.io \\
\thanks{\IEEEauthorrefmark{1} These authors contributed equally to this work}}}


%


\maketitle
\thispagestyle{fancy}
\begin{abstract}
This paper presents models for detecting agreement/disagreement in online discussions. In this work we show that by using a Siamese inspired architecture to encode the discussions, we no longer need to rely on hand-crafted features to exploit the meta thread structure. We evaluate our model on existing online discussion corpora – ABCD, IAC and AWTP. Experimental results on ABCD dataset show that by fusing lexical and word embedding features, our model achieves the state of the art performance of 0.804 average F1 score.  We also show that the model trained on ABCD dataset performs competitively on relatively smaller annotated datasets (IAC and AWTP).
\end{abstract}


%
\IEEEpeerreviewmaketitle

\section{Introduction}
The rise of various discussion forums and social media websites has given people a lot of avenues to express their opinions. As multiple people join a particular discussion, participants often agree or disagree with views presented by others. Mining the agreement and disagreement  (denoted (dis)agreement) signals helps detect presence of disputes, ideological stance of the participants\cite{somasundaran2009recognizing} and unravel beliefs shaping the opinion in general. This can further be useful for detecting subgroups\cite{hassan2012detecting}\cite{abu2012subgroup}, analyzing how well a new product is being received or analyzing the mood to predict the trends on stock markets\cite{bollen2011twitter}.

In this work, we explore a Siamese\cite{bromley1994signature} inspired deep neural network to detect the presence of (dis)agreement in online discussions between two posts, the quote and the response (Q-R pairs \cite{walker2012corpus}). In this framework, the same neural network encoder is applied to two input sentences individually, so that both of the two sentences are encoded into sentence vectors in the same embedding space. Prior work in this problem primarily focused on using handcrafted features to exploit the meta thread structure. We show that by training on a sufficiently large dataset (ABCD) we can bypass the need for designing handcrafted features. Thus, the classifier can be used for (dis)agreement detection between any two posts, even when the underlying hierarchical relationship between the Q-R pairs isn't available. To the best of our knowledge, this is the first work to investigate detection of (dis)agreement using sentence based encoding.

We detect (dis)agreement by performing a 3-way classification (agreement/disagreement/none) between the Q-R pairs on several existing annotated datasets. Due to the lack of a standard dataset, some prior work focused primarily on 2-way classification (agreement/disagreement).

In the following sections, we first discuss related work. Section \ref{data} describes the datasets used for evaluation. Section \ref{feature} describes the various features used in the classifier. In section \ref{system}, we explain the model architecture. Section \ref{results} details the experiments performed and their corresponding results. Section \ref{analysis} performs error analysis on the results from the proposed system. Finally, we conclude in section \ref{conclusion} and suggest relevant future work.

\section{Related Work} \label{related}
Previous work in this field focused a lot on spoken dialogues. \cite{hillard2003detection, galley2004identifying,hahn2006agreement} used spurt level agreement annotations from the ICSI corpus \cite{janin2003icsi}. \cite{germesin2009agreement} presents detection of agreements in multi-party conversations using the AMI meeting corpus \cite{mccowan2005ami}. \cite{wang2011detection} presents a conditional random field based approach for detecting agreement/disagreement between speakers in English broadcast conversations

Recently, researchers have turned their attention towards 
(dis)agreement detection in online discussions. The prior work was geared towards performing 2-way classification of agreement/disagreement. \cite{yin2012unifying} used various  sentiment, emotional and durational features to detect local and global (dis)agreement in discussion forums. \cite{abbott2011can} performed (dis)agreement on annotated posts from the Internet Argument Corpus (IAC) \cite{walker2012corpus}. They investigated various manual labelled features, which are however difficult to reproduce as they are not annotated in other datasets. To benchmark the results, we've also incorporated the IAC corpus in our experiments. Quite recently, \cite{rosenthal2015couldn} proposed a 3-way classification by exploiting meta-thread structures and accommodation between participants. They also proposed a naturally occurring dataset ABCD (Agreement by Create Debaters) which was about 25 times larger than prior existing corpus. We've trained our classifier on this larger dataset. \cite{wang2016improving} proposed (dis)agreement detection with an isotonic Conditional Random Fields (isotonic CRF) based sequential model. \cite{misra2013topic} proposed features motivated  by theoretical predictions to perform (dis)agreement detection. However, they've used hand-crafted patterns as features and these features miss few real world scenarios reducing the performance of the classifier.

(Dis)agreement detection is related to other similar NLP tasks like stance detection and argument mining but is not exactly the same. Stance detection is the task of identifying whether the author of the text is in favor or against or neutral towards a target, while argument mining focuses on tasks like automatic extraction of arguments from free text, argument proposition classification and argumentative parsing \cite{moens2007automatic} \cite{palau2009argumentation}. Recently there are studies on how people back up their stances when arguing where comments are classified as either attacking or supporting a set of pre-defined arguments \cite{boltuzic2014back}. These tasks (stance detection, argument mining) are not independent but have some common features because of which they are benefited by common building blocks like sentiment detection, textual entailment and sentence similarity \cite{boltuzic2014back} \cite{mohammad2017stance}.

\section{Data} \label{data}
In this work, we focus on 3-way classification (agreement/disagreement/none) between quote-response (Q-R) pairs for 3 prior existing in-domain datasets. These are described in the subsequent sub-sections.

\subsection{Agreement by Create Debaters (ABCD)} 
The ABCD corpus \cite{rosenthal2015couldn} was curated from Create Debate website\footnote{\url{http://www.createdebate.com/}} where users can start a debate by asking a question. Although the website can support open ended as well as multiple sided debates, the corpus comprises only of the for-against debates. The corpus is annotated as follows: the side label corresponding to each post (\textit{response}) determines whether the user agrees or disagrees with the previous post (\textit{quote}). If the authors of both the posts are different, then they agree if the side labels are same or otherwise disagree. If the authors of both the posts is same, it is labeled as none as it implies that it is in continuation of the previous post. Also, the first post in a debate is usually setting up the premise of the debate, so it doesn't have a side attached to it. Hence all the Q-R pairs with the quote as the first post are labeled as none. Table \ref{abcd_examples} shows example Q-R pairs for each label type.

\begin{table}[!htbp]
\caption{Examples of agreement/disagreement/none in Quote-Response pairs in ABCD dataset}
\label{abcd_examples}
\centering

\begin{tabular}{|c|c|}\hline
\textbf{Label} & \textbf{Post} \\ \hline
\multirow{ 2}{*}{\textbf{None}}  & \multicolumn{1}{m{7cm}|}{\textit{Quote}: Is Scientology a real religeon? Or is it a fake money making gimmick?} \\\cline{2-0}
 & \multicolumn{1}{m{7cm}|}{\textit{Response}: All religions are fake, there's an argument to be made the vast majority are money making gimmicks. Scientology is no more outlandish than any of the more widespread religions.}\\\hline \hline

\multirow{ 2}{*}{\textbf{Agree}}  & \multicolumn{1}{m{7cm}|}{\textit{Quote}: I am against suicide because you are basically not only harming yourself, but everyone else around you. Let's not mention it is a cowards way out. I also have a religious but I CAN explain that reason.} \\\cline{2-0}
 & \multicolumn{1}{m{7cm}|}{\textit{Response}: So true man people only harm the people they love by dying. I am not religious and religiously and non-religiously suicide is wrong.}\\\hline \hline

\multirow{ 2}{*}{\textbf{Disagree}} & \multicolumn{1}{m{7cm}|}{\textit{Quote}: The majority of the information learned in school is irrelevant to real world skills. Besides, in a voluntary setting, most children would go to school via parents demands where school choice would be much more abundant.} \\\cline{2-0}
 & \multicolumn{1}{m{7cm}|}{\textit{Response}: Children learn math which is relevant, children learn history which is relvant, children learn the releveant languge to their country, children learn foreign languages which imporoves economic opportunites. My one friend grew up in Baghdad, Iraq, and they don't play when it comes to education. He started learning English in the 3rd grade I think through graduation which helped his economic ooportunities, and he is an artchitect so the math helped. Please excuse my typos. I have a learning disability.}\\\hline

\end{tabular}
\end{table}

\subsection{Agreement in Wikipedia Talk Pages (AWTP)}
AWTP \cite{andreas2012annotating} is formatted in the same way as ABCD. Post-reply pairs are manually annotated with their (dis)agreement stance. Also additional mode information indicates the manner in which agreement or disagreement is expressed. The datasource for AWTP comprised primarily of Wikipedia Talk Pages and LiveJournal postings. Table \ref{awtp_stats} provides additional statistics for the corpus. 

\begin{table}[!htbp]
\caption{Unique annotation counts for AWTP dataset}
\label{awtp_stats}
\centering
\begin{tabular}{|c||c|c|c||c|c|c|}
\hline & \multicolumn{3}{|c||}{\textbf{Wikipedia}} & \multicolumn{3}{|c|}{\textbf{LiveJournal}} \\
\hline
 & Agree & Disagree & None & Agree & Disagree & None\\
\hline
\textbf{Train} & 219 & 471 & 703 & 390 & 83 & 0\\
\hline
\textbf{Dev} & 69 & 101 & 24 & 0 & 0 & 0\\
\hline
\textbf{Test} & 62 & 107 & 79 & 0 & 0 & 0\\
\hline
\end{tabular}
\end{table}

\subsection{Internet Argument Corpus (IAC)}
The Internet Argument Corpus (IAC) \cite{walker2012corpus} is a collection of corpora for research in political debate on internet forums. It consists of  \url{~}11,000 discussions, \url{~}390,000 posts, and some \url{~}73,000,000 words. It includes topic annotations, response characterizations, and stance. The 4forum posts were annotated using Mechanical Turk. The annotators were provided with a Q-R pair and they indicated the level of (dis)agreement on a scale of [-5, 5]. However, not all posts in a thread
were annotated for (dis)agreement and roughly 6000 valid Q-R pairs were extracted. In accordance with prior work of this corpus \cite{abbott2011can, misra2013topic, rosenthal2015couldn}, we converted the scalar values into corresponding (dis)agreement as follows: [-5, -1] is tagged as disagreement, [-1, 1] is tagged as none, [1, 5] is tagged as agreement. In case multiple annotators have tagged the same post, we combine them as follows. None annotations are ignored unless there are no other (dis)agreement tags. In all other cases, average annotation score is used as the final score of the post.

\section{Feature Extraction} \label{feature}
In this section we briefly mention the features experimented in this work.

\subsection{Word Vectors}
In the recent times distributed representations
of words \cite{turian2010word} (word2vec\cite{mikolov2013distributed}, GloVe\cite{pennington2014glove}) has shown promise in many NLP tasks and is the driver for success of deep learning in NLP \cite{karpathy2015deep}. Word vectors encode semantics in low dimensional space and can be used efficiently for various NLP tasks \cite{levy2014linguistic} \cite{levy2015improving}. For this task of (dis)agreement classification, we use GloVe embeddings of 300 dimensions trained on Common Crawl with 840 billion tokens, 2.2 million vocabulary.

\subsection{Lexicons}\label{lexicons}
We used affect, sentiment, emotion, opinion lexicons for feature extraction because in many of the online discussions forums people tend to argue with emotion and opinion about a particular topic to convey their stance or belief. Making use of these affect lexicons will help us in classifying if a response (dis)agrees with quote or not. Prior work\cite{yin2012unifying} using these lexicons have shown them to give good results. Among lexical features we used the following.

AFINN\cite{nielsen2011new} word list are manually rated for valence with an integer between -5 (Negative Sentiment) and +5 (Positive Sentiment). Bing Liu\cite{hu2004mining} opinion lexicon extract opinion on customer reviews. +/-EffectWordNet\cite{choi2014+} by MPQA group are sense level lexicons. The NRC Affect Intensity\cite{mohammad2017word} lexicons provide real valued affect intensity. NRC Word-Emotion Association Lexicon\cite{mohammad2010emotions} contains 8 sense level associations (anger, fear, anticipation, trust, surprise, sadness, joy, and disgust) and 2 sentiment level associations (negative and positive). Expanded NRC Word-Emotion Association Lexicon\cite{bravo2016determining} expands the NRC word-emotion association lexicon for twitter specific language. NRC Hashtag Emotion Lexicon\cite{mohammad2015using} contains emotion word associations computed on emotion labeled twitter corpus via Hashtags. NRC Hashtag Sentiment Lexicon and Sentiment140 Lexicon\cite{mohammad2013nrc} contains sentiment word associations computed on twitter corpus via Hashtags and Emoticons. SentiWordNet\cite{baccianella2010sentiwordnet} assigns to each synset of WordNet three sentiment scores: positivity, negativity, objectivity. Negation lexicons collections are used to count the total occurrence of negative words. The Linguistic Inquiry Word Count (LIWC)\cite{tausczik2010psychological} categorizes the words we use in everyday language to reveal our thoughts, feelings, personality, and motivations.

For each word in the sentence we calculate various metrics using these lexicons like number of negation words in a sentence, average negative and positive sentiment of the sentence etc. and use these as lexical feature vector to the system. The lexicon feature extractor was inspired from  \cite{duppada2017wassa}.

\section{System Description} \label{system}
The task of (dis)agreement classification from Q-R pair comes under a much broader category of sentence pair modelling. A lot of NLP tasks like natural language inference, textual entailment, answer selection, paraphrase identification etc involve modelling a pair of sentences so that they perform well on a particular task or multitude of such tasks\cite{yin2015abcnn}. For this task we used an architecture which is inspired by Siamese network\cite{bromley1994signature} and Stanford Natural Language Inference model\cite{bowman2015large} with identical networks encoding Q-R pair.

\begin{figure}[!htbp]
    \centering 
    \includegraphics[width=\columnwidth]{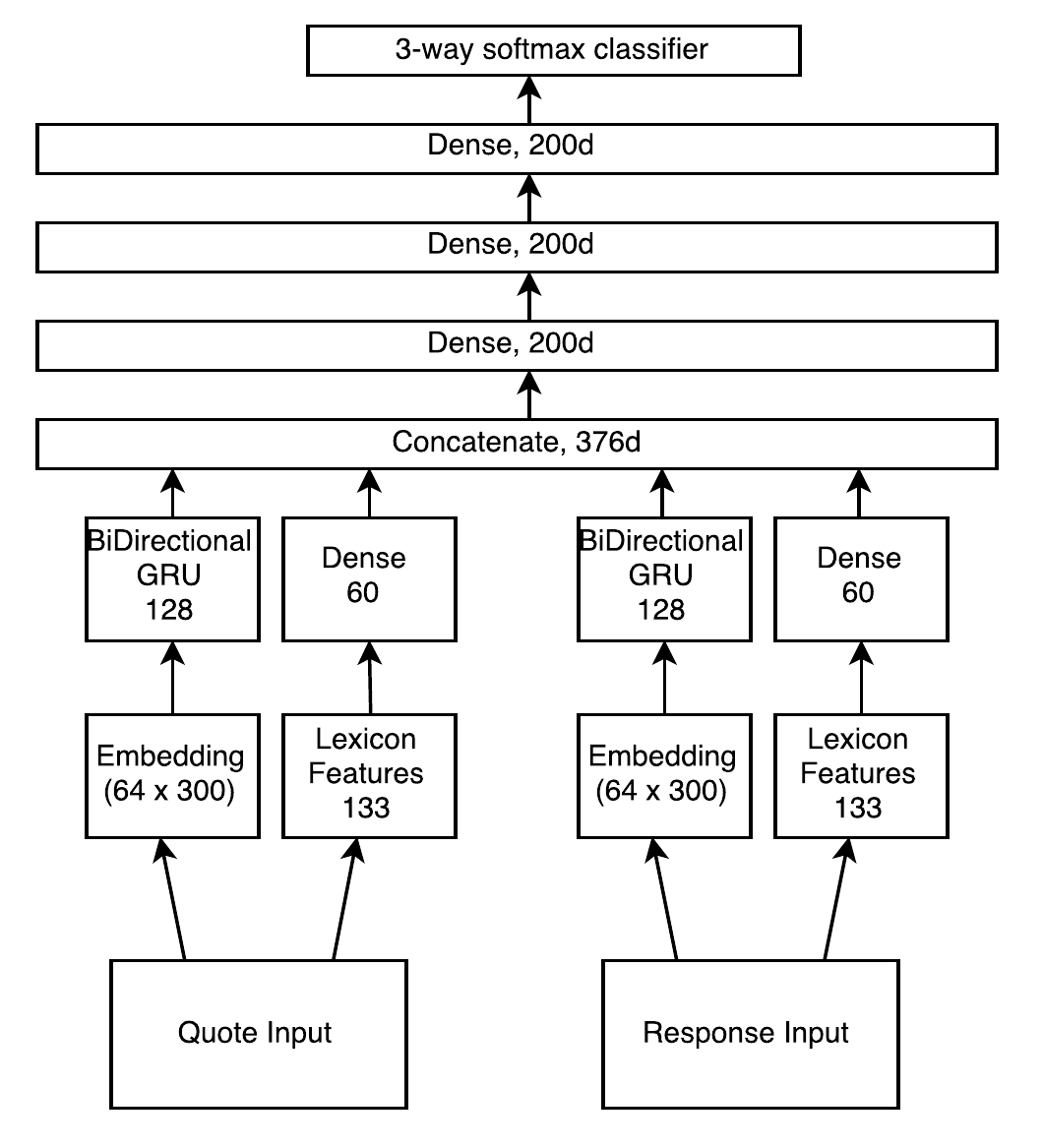}
	\caption{System Architecture}
	\label{fig:system_description}
\end{figure}

For each Q-R pair we extract two sets of features. First, GloVe word embeddings are fed to Gated Recurrent Units\cite{chung2014empirical} to create a sentence embedding. Second, from each text a lexical feature vector is extracted as mentioned in sub-section \ref{lexicons}. Both these sentence embeddings from Q-R pairs are concatenated and then fed into fully connected layers to do 3 way classification. Figure \ref{fig:system_description} show the architecture of our model.

\subsection{System Parameters} \label{params}
\begin{figure}[!htbp]
    \centering 
    \includegraphics[width=\columnwidth]{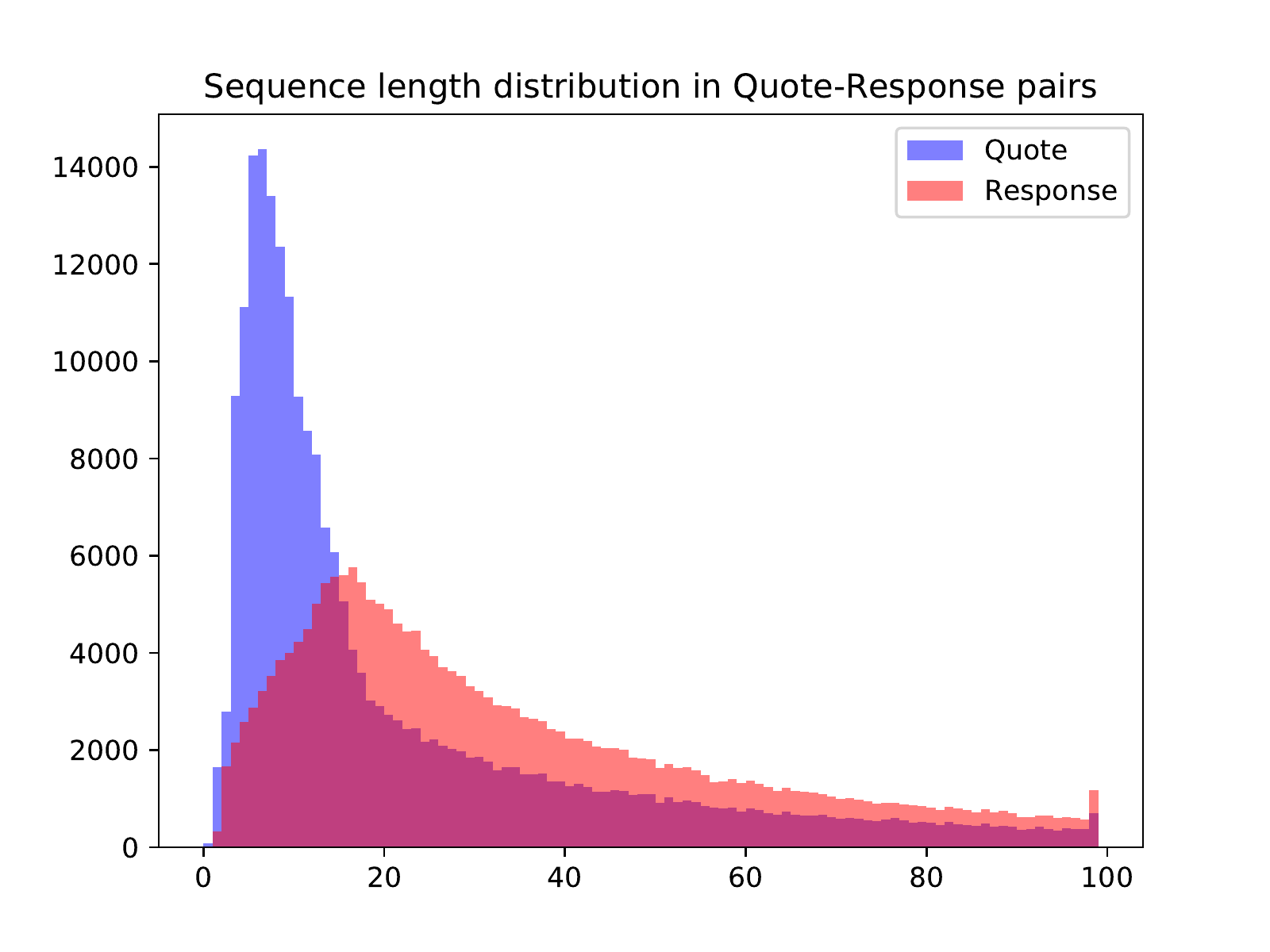}
	\caption{Distribution of sequence length in Q-R pairs. The graph shows the number of posts v/s sequence length.}
	\label{fig:seqlen}
\end{figure}

We have used relu non-linearity, dropout\cite{srivastava2014dropout} for regularization and batch normalization\cite{ioffe2015batch} for accelerating training. The network is optimized with adam\cite{kingma2014adam} optimizer with learning rate of 0.001. We have plotted sequence length of Q-R pairs in Figure \ref{fig:seqlen} to better estimate the maximum sequence length for GRU layer. The average sequence length of quotes and responses is 46.28 and 64.935 respectively. The Impact of maximum sequence length of the text input to GRUs is computed and can be found in  section \ref{results}. We used keras\cite{chollet2015keras} deep learning framework to train our models.

\section{Experiments} \label{results}
In this section, we evaluate our model on 3-way (dis)agreement classification. The general settings of the model have been defined in sub-section \ref{params}. In the upcoming sub-sections, we explore the variants of our model and compare our model with the state of the art models on benchmark datasets from section \ref{data}.

\subsection{Features Used}
We implemented three variants of the architecture: only lexical features are used, only GloVe embeddings are used and finally where both of them are used. As is evident and hypothesized, the model trained with lexical and word embeddings gave the best results. It is interesting to see that using just the lexicons, the model beat the previous best model which signifies the importance of lexical features. However, using only lexicons as features will suffer from all the disadvantages of any bag-of-words model. This is primarily because it cannot encode the temporal nature of language. This is where gated recurrent units or in general recurrent neural networks come into play. GRU's can successfully encode the temporal nature. Thus, by fusing both the GRU encoded embeddings and the lexicons, we achieved the state of the art results on the ABCD dataset by beating the previous best model's \cite{rosenthal2015couldn} average F1 score with margin of 4\% percentage. Table \ref{abcd_results_features} compares variants of the proposed system with the existing state of the art.

\begin{table}[!htbp]
    \caption{Analyzing the impact of changing the feature vector on ABCD corpus. Entries marked with \textbf{bold} are the best performing variants and those marked with - are not available.}
    \label{abcd_results_features}
	\centering
	\begin{tabular}{|c|c|c|c|}
		\hline
		\textbf{System} & \textbf{Precision} & \textbf{Recall}  & \textbf{Weighted F1 Score} \\
		\hline
		\textbf{SOTA \cite{rosenthal2015couldn}} & 0.776 & - & - \\
		\textbf{Lexicons} & 0.788 & 0.798  & 0.789 \\
		\textbf{GRU} & 0.792 & 0.798 & 0.794 \\
		\textbf{GRU + Lexicons} & \textbf{0.812} & \textbf{0.815} & \textbf{0.804} \\
		\hline
	\end{tabular}
\end{table}

\subsection{Maximum Input Sequence Length}
We investigate the impact of varying the maximum input sequence length in Table \ref{abcd_results_seq}. The results highlight some interesting insights. Since we are dealing with discussion posts, the average post size is larger than a few sentences. Hence, as we increase the maximum sequence length, the increase in performance is justified. However, once we increase the size after a certain threshold, a lot of the input sequences need to be padded with zeroes, thus causing a drop in performance.

\begin{table}
	\caption{Investigating the impact of varying the maximum sequence length on the overall model performance. Entries marked with \textbf{bold} are the best performing variants}
	\label{abcd_results_seq}
	\centering
	\begin{tabular}{|c|c|c|c|}
		\hline
		\textbf{Sequence Length} & \textbf{Precision} & \textbf{Recall}  & \textbf{Weighted F1 Score} \\
		\hline
		\textbf{32} & 0.810 & 0.813 & \textbf{0.806} \\
		\textbf{64} & \textbf{0.812} & \textbf{0.815} & 0.804 \\
		\textbf{128} & 0.805 & 0.808 & 0.796 \\ 
		\hline
	\end{tabular}
\end{table}

\subsection{Transfer Learning} \label{transfer_learning}
In the recent times, with the availability of huge amount of data a new paradigm in machine learning called Transfer Learning\cite{pan2010survey} has come into play. Transfer learning is the improvement of learning in a new task through the transfer of knowledge from a related task that has already been learned. Here we've applied transfer learning technique to smaller datasets and achieved competitive results.
Table \ref{iac_results} and \ref{awtp_results} enlist the results of various model architectures explored for testing the effectiveness of transferring learning from ABCD dataset to the smaller annotated datasets IAC and AWTP.
The variants explored are as follows. \textit{Direct}, where the model trained on ABCD model (referred as pre-trained model) is tested directly on the smaller dataset. \textit{Tuning}: the model is seeded with the weights from the pre-trained model and trained with the smaller dataset. \textit{Transfer}: The last 2 layers from the pre-trained model are stripped and replaced with new dense layers of size 100 and 50 and the model is trained on the smaller dataset. \textit{Re-train last 2/3 layers}: All but last-2/3 layers of the pre-trained model are frozen and the remaining layers are trained on the smaller dataset. 

\begin{table}
    \caption{Tuning model trained on ABCD dataset for IAC dataset. Entries in \textbf{bold} are the best performing variants and those marked with - are not available. The model variants are explained in sub-section \ref{transfer_learning}}
    \label{iac_results}
	\centering
	\begin{tabular}{|c|c|c|c|}
		\hline
		\textbf{Model} & \textbf{Precision} & \textbf{Recall}  & \textbf{Weighted F1 Score} \\
		\hline
		\textbf{SOTA \cite{rosenthal2015couldn}} &  - & - & \textbf{0.578} \\
		\textbf{Direct} &  0.473 & 0.364 & 0.285 \\
		\textbf{Tuning} &  0.530 & \textbf{0.572} & 0.450 \\
		\textbf{Transfer} &  0.508 & 0.559 & 0.465 \\
		\textbf{Re-train last 2 layers} &  \textbf{0.531} & \textbf{0.572} & 0.460 \\
		\textbf{Re-train last 3 layers} &  0.523 & 0.570 & 0.428 \\
		\hline
	\end{tabular}
\end{table}

\begin{table}
    \caption{Tuning model trained on ABCD dataset for AWTP dataset. Entries in \textbf{bold} are the best performing variants and those marked with - are not available. The model variants are explained in sub-section \ref{transfer_learning}}
    \label{awtp_results}
	\centering
	\begin{tabular}{|c|c|c|c|}
		\hline
		\textbf{Model} & \textbf{Precision} & \textbf{Recall}  & \textbf{Weighted F1 Score} \\
		\hline
		\textbf{SOTA \cite{rosenthal2015couldn}} &  - & - & 0.389 \\
		\textbf{Direct} &  0.434 & 0.445 & 0.389 \\
		\textbf{Tuning} &  0.515 & 0.470 & 0.464 \\
		\textbf{Transfer} &  0.525 & 0.437 & 0.412 \\
		\textbf{Re-train last 2 layers} &  \textbf{0.534} & 0.457 & 0.447 \\
		\textbf{Re-train last 3 layers} &  0.477 & \textbf{0.546} & \textbf{0.486} \\
		
		\hline
	\end{tabular}
\end{table}

\section{Error Analysis} \label{analysis}
The ABCD dataset is scraped from online debate forum, Create Debate and is automatically labelled. This way of collecting Q-R pairs is not perfect and suffers from the following problems: people may be on the same/different side of debate but disagree/agree on some points (example 1 in Table \ref{bad_eggs}) as the sides are for topic level not post or sentence level, off-topic social chitter-chatter on debate forum (example 2 in Table \ref{bad_eggs}) etc. Our analysis indicated that in quite a few cases, the error in classification was in fact a case of incorrect label. 

\begin{table}[!htbp]
\caption{Hard Examples of ABCD Dataset}
\label{bad_eggs}
\centering

\begin{tabular}{|c|c|}\hline
\textbf{Label} & \textbf{Post} \\ \hline
 
 \multirow{ 2}{*}{\textbf{agree}}  & \multicolumn{1}{m{7cm}|}{\textit{Quote}: JessHall01 my parents do not deserve it, they treat me like sh and that's unecessary. NO parents should even romotely lay their hands on a child. EVER.} \\\cline{2-0}
 & \multicolumn{1}{m{7cm}|}{\textit{Response}: kamranw I agree with you there. NO parent should hit their child. I completely feel your pain if that is the case. I would be curious to know how they treat you like shit though. Not sure how that is relevent to minors being sexual active either.}\\\hline \hline
 
 \multirow{ 2}{*}{\textbf{disagree}}  & \multicolumn{1}{m{7cm}|}{\textit{Quote}: joecavalry I tell my kids to hurry up and eat their breakfast so that we can get to school on time. My little one is so slow that I tell her she eats breakslow ;)} \\\cline{2-0}
 & \multicolumn{1}{m{7cm}|}{\textit{Response}: Morgie7171 AWWW thats so cute............... hahahahahahahahahahhaahahhha}\\\hline
 
\end{tabular}
\end{table}

\section{Conclusion} \label{conclusion}
We have trained a deep neural network with fusion of lexical and word vector based features to achieve state of the art results on 3-way (dis)agreement classification on the largest (dis)agreement  classification dataset available till date (ABCD corpus). We've shown that by using this model, we no longer need to rely on hand-crafted features to exploit the meta-thread structure. We've also shown the benefit of transfer learning from pre-trained model on large corpora to achieve competitive results on small domain datasets. Till date, the research for (dis)agreement classification has not moved at the pace comparable to some other NLP tasks primarily because of unavailability of large standard dataset. Although ABCD dataset is large enough it suffers from many issues as it uses naturally occuring labels. We plan to enhance the ABCD dataset by performing semi-supervised tagging of labels by training models on hand-annotated datasets. With the availability of a large standard dataset for (dis)agreement classification, we can try various recent advanced state of the art architectures for modelling sentences pairs\cite{yin2015abcnn}, \cite{wang2017bilateral} to further improve the performance.



%
\bibliographystyle{IEEEtran}
\bibliography{IEEEreference}

\end{document}